\documentclass[10pt,twocolumn,letterpaper]{article}

\usepackage{iccv}
\usepackage{times}
\usepackage{epsfig}
\usepackage{graphicx}
\usepackage{amsmath}
\usepackage{amssymb}

\usepackage{enumitem}
\usepackage{multicol}
\usepackage{bm}
\usepackage[titlenumbered,ruled, vlined]{algorithm2e}
\usepackage{algorithmic}
\usepackage{ragged2e}
\usepackage{caption}
\usepackage{xcolor}
\usepackage[para,online,flushleft]{threeparttable}
\usepackage{booktabs}
\usepackage{multicol}
\usepackage{multirow}
\usepackage{subfigure}
\usepackage[pagebackref=true,breaklinks=true,colorlinks,bookmarks=false]{hyperref}
\usepackage{xpatch}
\pdfoutput=1

\DeclareMathOperator*{\argmin}{argmin}

\iccvfinalcopy 

\def\iccvPaperID{8642} 

\newcommand{\formattedparagraph}[1]{\noindent \textbf{#1}}

\ificcvfinal\fi

\begin{document}

\title{Trilevel Neural Architecture Search for Efficient Single Image Super-Resolution}

\author{Yan Wu\textsuperscript{\rm 1}, Zhiwu Huang\textsuperscript{\rm 1}, Suryansh Kumar\textsuperscript{\rm 1}, Rhea Sanjay Sukthanker\textsuperscript{\rm 1}, Radu Timofte\textsuperscript{\rm 1}, Luc Van Gool\textsuperscript{\rm 1,2}\\
	\textsuperscript{\rm 1}Computer Vision Lab, ETH Z\"urich, Switzerland \quad
	\textsuperscript{\rm 2}PSI, KU Leuven, Belgium\\
	{\tt\small wuyan@student.ethz.ch, rhea.sukthanker@inf.ethz.ch,}\\ {\tt\small\{zhiwu.huang, sukumar, radu.timofte, vangool\}@vision.ee.ethz.ch}
}

\maketitle
\ificcvfinal\thispagestyle{empty}\fi

\begin{abstract}

Modern solutions to the single image super-resolution (SISR) problem using deep neural networks aim not only at better performance accuracy but also at a lighter and computationally efficient model. To that end, recently, neural architecture search (NAS) approaches have shown some tremendous potential.
Following the same underlying, in this paper, we suggest a novel trilevel NAS method that provides a better balance between different efficiency metrics and performance to solve SISR. Unlike available NAS, our search is more complete, and therefore it leads to an efficient, optimized, and compressed architecture. We innovatively introduce a trilevel search space modeling, i.e., hierarchical modeling on network-, cell-, and kernel-level structures. To make the search on trilevel spaces differentiable and efficient, we exploit a new sparsestmax technique that is excellent at generating sparse distributions of individual neural architecture candidates so that they can be better disentangled for the final selection from the enlarged search space. We further introduce the sorting technique to the sparsestmax relaxation for better network-level compression. 
The proposed NAS optimization additionally facilitates simultaneous search and training in a single phase, reducing search time and train time. Comprehensive evaluations on the benchmark datasets show our method's clear superiority over the state-of-the-art NAS in terms of a good trade-off between model size, performance, and efficiency.

\end{abstract}
\section{Introduction}

Single image super-resolution (SISR) aims at upsampling a single low-resolution image to its high-resolution description.  It is well-known that image SR is an ill-posed problem, and lately, many deep learning-based methods have emerged to address it~\cite{SRCNN,  DRCN,  ESRGAN, SAN, SRDenseNet, SRFBN, SRGAN, lee2020journey}. Although learning-based methods achieve good performance, the computational and memory requirement greatly affects its on-device usage. Motivated by this, our paper focuses on learning efficient SISR with neural architecture search (NAS) methods. Our method aims to keep a good performance-efficiency balance among peak signal-to-noise ratio (PSNR),  floating-point operations per second (FLOPS), parameter size, number of employed GPUs, and search time of neural architectures.

\begin{figure}[t]
\begin{center}
   \includegraphics[width=0.55\linewidth]{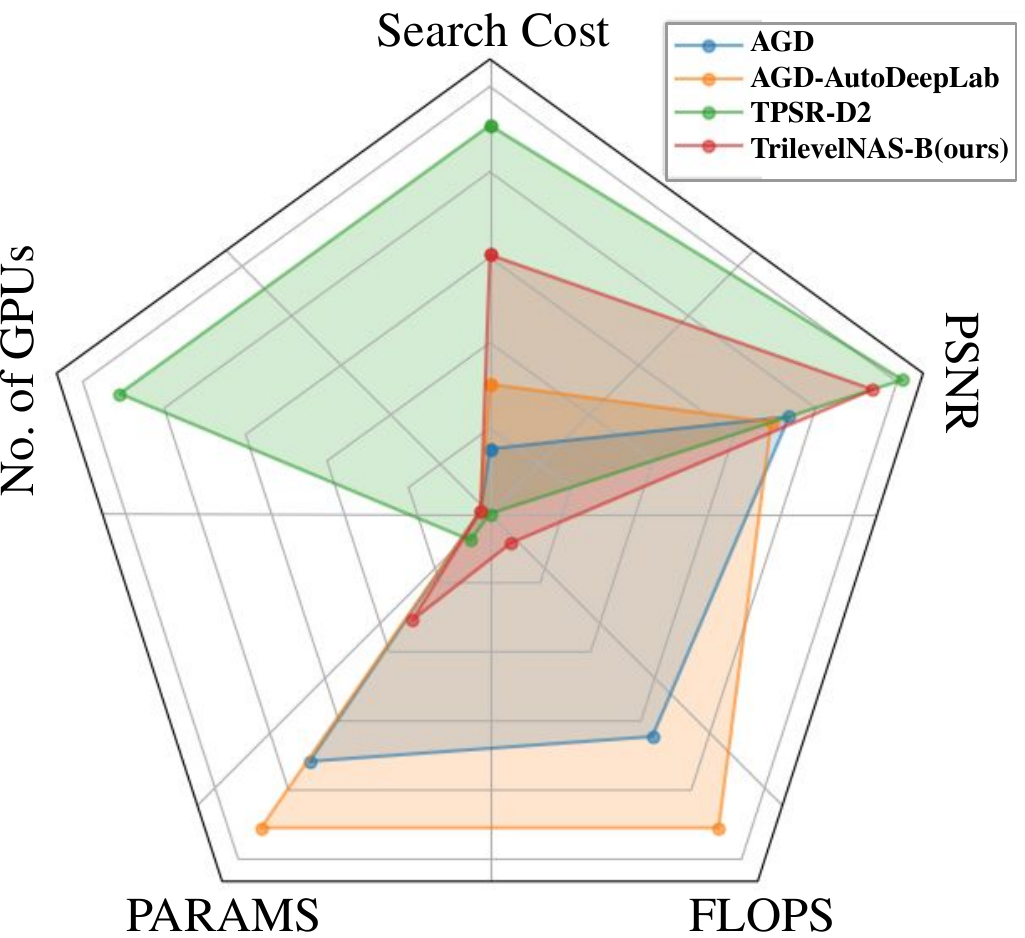}
\end{center}
   \caption{\footnotesize Radar plots (\emph{smaller covering area shows better method}) of the proposed TrilevelNAS and three state-of-the-art NAS-based SR methods, i.e., AutoGAN-Distiller (AGD) \cite{AGD}, AutoDeepLab \cite{Autodeeplab} and tiny perceptual SR (TPSR) \cite{lee2020journey}, in terms of our focussed five metrics: inverse PSNR, FLOPS, parameter size, number of used GPUs, and search time. The smallest covering area shows that our TrilevelNAS method work clearly better (i.e., keep the best balance among the five metrics) than the others.}
\label{fig:chart_comparison}
\vspace{-0.3cm}
\end{figure}

\begin{figure}[t]
\begin{center}
   \includegraphics[width=0.65\linewidth]{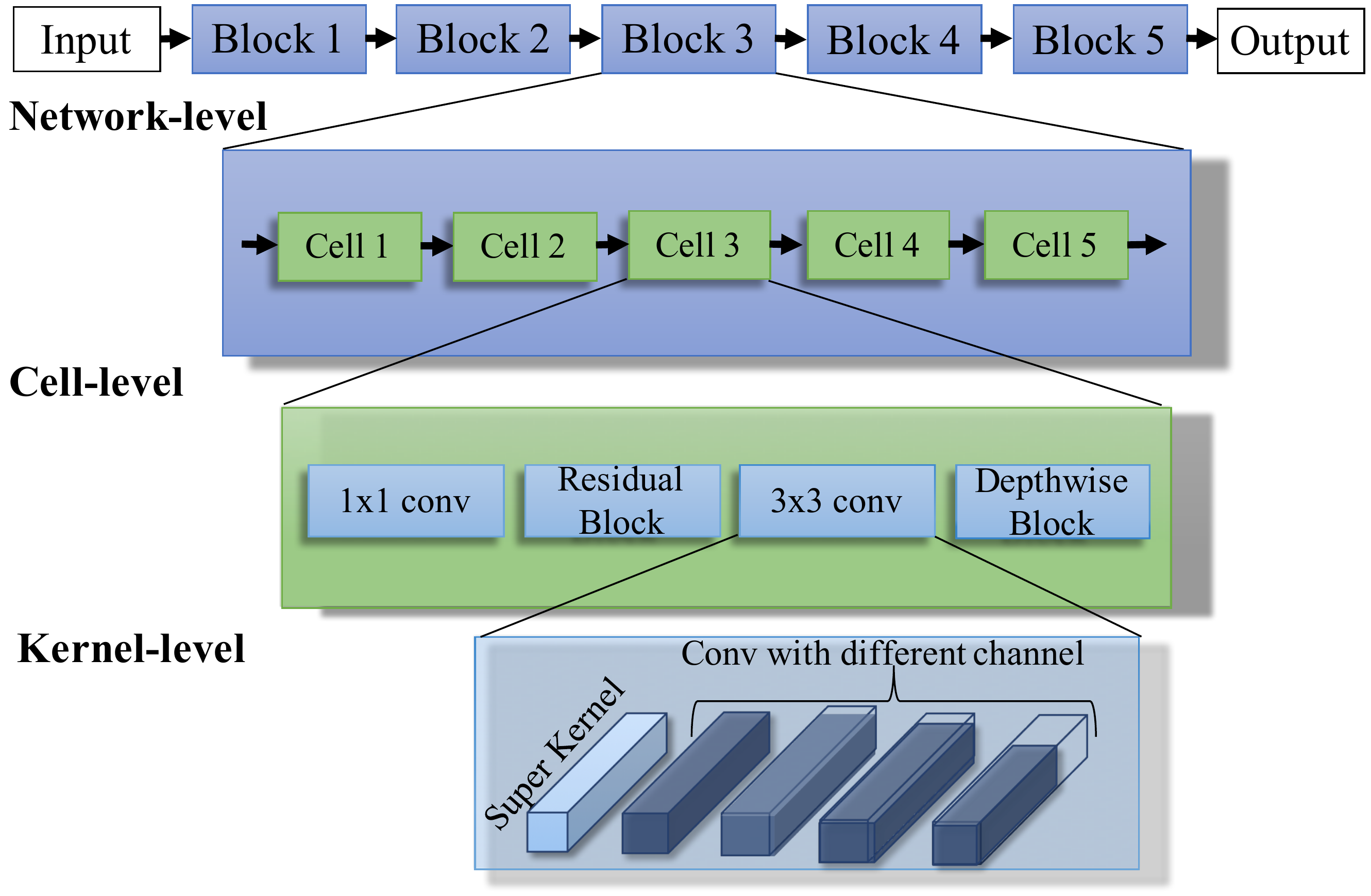}
\end{center}
   \caption{\footnotesize A typical trilevel neural architecture design for SISR.}
\label{fig:TrilevelNAS_overview}
\vspace{-0.4cm}
\end{figure}

As illustrated in Fig.~\ref{fig:chart_comparison}, there have been emerging a few NAS-based methods for efficient SISR. Generally, they aim at searching for one or two of the following three-level neural architectures, \ie, \emph{(i)} A \textbf{network-level} optimization to properly position the upsampling layers \cite{Autodeeplab, Guo_2020}. \emph{(ii)} A \textbf{cell-level} optimization to improve the capacities of encoding or upscaling \cite{DARTS, chu2020fair}. \emph{(iii)} A \textbf{kernel-level} optimization to make a trade-off between an operator's capacity and its width (\ie, the number of input/output channels) for a compressed model \cite{9157494, AGD} (see Fig.\ref{fig:TrilevelNAS_overview}). In particular,  AutoGAN-Distiller (AGD) \cite{AGD} exploits a differentiable NAS method to search for optimal architectures on both cell- and kernel-levels. As this method overlooks the network-level compression, the learned neural architectures are still of big model size and slow FLOPS. AutoDeepLab \cite{Autodeeplab} proposes a differentiable NAS algorithm to seek effective network- and cell-level architectures. Missing kernel-level search makes this method still suffer from unsatisfactory model compression and FLOPS reduction. In contrast, the tiny perceptual SR (TPSR) method \cite{lee2020journey} exploits a reinforcement learning (RL) based NAS method to search for optimal cell-level architecture so that a tiny SR architecture is finally discovered. Like most RL-based NAS algorithms, success of TPSR requires extortionate search cost: 40 NVIDIA GeForce GTX1080 Ti server GPUs with 12 days of search time.

To overcome all the shortcomings mentioned before, this paper introduces a new trilevel NAS algorithm. Our method efficiently solves the SR problem with the on-device computational resource\footnote{By on-device, we mean a computer with 1 GPU (16GB RAM).}, and gives performance accuracy comparable to the best available methods (see Fig. \ref{fig:chart_comparison}). It automates the whole trilevel neural architectures' design optimization and does not explicitly use ground-truth examples for supervision. For that, we first define the discrete search space at different levels and its modeling as follows.

\smallskip
\formattedparagraph{(a) Search Space.}
We define search space at three different levels (see Fig.\ref{fig:TrilevelNAS_overview}). The network-level search space is composed of all the candidate network paths. The cell-level search space contains a mixture of all possible candidate operations. Lastly, the kernel-level search space is a single convolutional kernel with a defined subset of the convolution kernel dimensions \S \ref{sec:search_space}. 

\begin{figure}[t]
\begin{center}
 \includegraphics[width=0.65\linewidth]{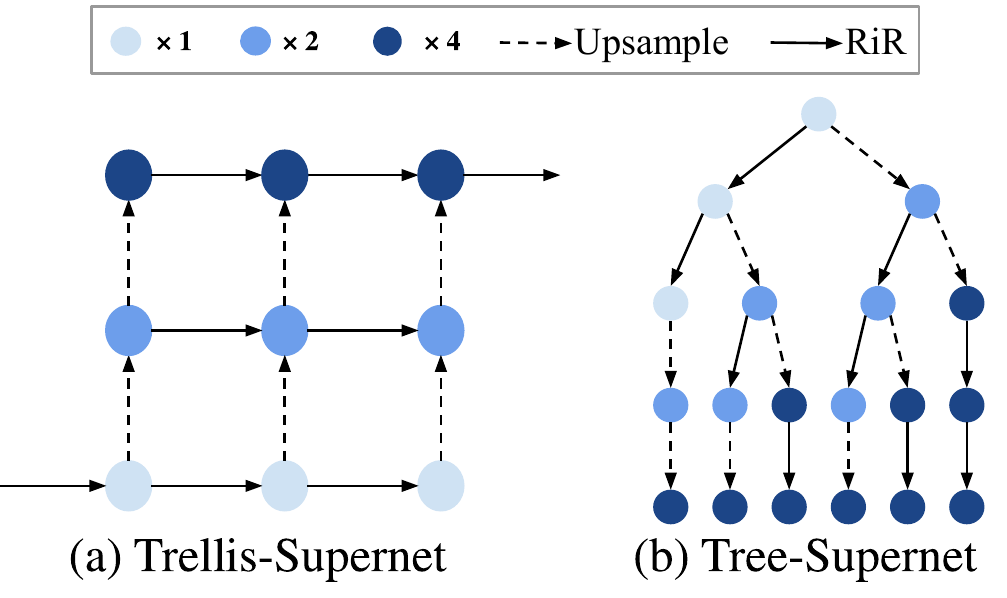}
\end{center}
\caption{\footnotesize (a) Trellis-supernet modeling \cite{Autodeeplab} on 2 Residual-in-Residual (RiR) blocks and 2 Upsampling blocks. Vertical and horizontal axes show upsampling scale and layer number, respectively. Each intermediate feature map (blue) has two paths to traverse: horizontal (RiR block) and vertical ($\times2$Upsampling layer). A feasible path traverse from the start of the node to the end node. (b) The proposed tree-supernet modeling with 2 RiR and 2 Upsampling layers. Like Trellis, each intermediate feature map has 2 paths. However, each path is independent and do not merge. Any path that starts from the root is a feasible path.}
\label{fig:tree-superpath}
\vspace{-0.4cm}
\end{figure}

\smallskip
\formattedparagraph{(b) Modeling of Search Space.} We perform continuous relaxation of the defined discrete search space to be optimized efficiently in a differentiable manner. We follow Differentiable Architecture Search (DARTS) \cite{DARTS} and AGD \cite{AGD} to model cell- and kernel-level search spaces with supercell and superkernel, respectively.
However, they both overlook the modeling of network-level search space, which is crucial for network-level optimization and compression.
One way is to adopt the trellis-like supernet modeling in Fig.\ref{fig:tree-superpath}(a) from AutoDeepLab~\cite{Autodeeplab}. Yet, in trellis-supernet, any possible path's information flow is highly-entangled, and thus the derived path and cell architectures may not be optimal. Also, we get a fixed network path length from it. To overcome those issues, we introduce a tree-like supernet, where paths are enumerated independently and hence can be pruned more easily once it makes a reasonable contribution to the supernet (see Fig.\ref{fig:tree-superpath}(b)). Intuitively, we can observe from Fig.\ref{fig:tree-superpath} that our tree-like supernet disentangles the sequential network path dependency and provides a better trade-off between memory consumption, cell sharing across layers, and search time.

For continuous relaxation of the search spaces, differentiable NAS methods often employ softmax \cite{DARTS}. But, softmax may not give a clear-cut dominant operation for selection to optimal architecture design.  Hence, we exploit the sparsestmax strategy \cite{sparsestmax} for continuous relaxation of cell-level search space. The sparsestmax generates sparse distribution and preserves softmax's vital properties (\eg, differentiability and convexity). For network-level search space, we exploit sorted sparsestmax, where the network level path sparsity is arranged in descending order. The proposed sorted sparsestmax helps to prune the tail of the network path for notable model compression. Our proposition to use sparsestmax and its variant for network-level modeling enables a new NAS training scheme. Using that, we can perform training from search rather than from scratch, as practiced by most NAS algorithms.

For evaluation and comparison of our method, we used standard Set5 \cite{bevilacqua2012low}, Set14 \cite{zeyde2010single}, and AIM 2020 SR dataset \cite{zhang2020aim}. In this paper, we make the following contributions.
\begin{itemize}[leftmargin=*,topsep=0pt, noitemsep]
    \item For more complete neural architecture optimization and compression, we propose a novel trilevel NAS algorithm to simultaneously search for network-, cell- and kernel-level neural architectures, which is rarely studied in the NAS literature. 
    
    \item  To make the search on the trilevel spaces differentiable and efficient, we exploit a new relaxation technique that is good at producing a better disentanglement among candidate neural architectures. Besides, it enables network-level compression as well as a continuous merge between the search and train phases.
    
    \item Compared to the state-of-the art methods, our method plays a good balance between performance and efficiency on Set5, Set14, and AIM 2020 datasets, without using ground-truth data during training for real-world SR tasks.
\end{itemize}

\section{Relevant Work}
Numerous solutions to the SISR problem based on one-level and bi-level NAS core have emerged, recently. We discuss them accordingly for better understanding.

\smallskip
\formattedparagraph{One-level NAS methods.} Evolutionary algorithms (e.g., \cite{real2019regularized, real2017large}), and reinforcement learning-based NAS methods (e.g., \cite{zoph2017neural, zoph2018learning, lee2020journey}) introduced input/output channel numbers into their search space for optimization. Zhang \etal ~\cite{zhang2020memoryefficient} suggested a differentiable width-level architecture search method. They aim to select an optimal policy by keeping the same kernel-width, half the kernel-width, and double the kernel-width at each layer. Recently, Lee \etal \cite{lee2020journey} proposed a tiny SR method that integrates reinforcement learning-based NAS and GANs but requires huge computational resources and search time. Another variant is Liu \etal~\cite{DARTS} DARTS, which models the basic cell of the network with a directed acyclic graph (DAG). It relaxes the given discrete search space into a continuous one via a softmax combination of all the candidate operations on each edge, constituting a supernet. Such relaxation enables differentiable optimization of the supernet. Though effective, it suffers from problems such as: large gap between the supernet and stand-alone architecture \cite{chen2020progressive}, aggregation of skip-connections \cite{chu2020fair} and large memory consumption \cite{proxylessnas}.

\smallskip
\formattedparagraph{Bi-level NAS methods.} Stamoulis \etal \cite{stamoulis2019single} encodes candidate cell-level operations and kernel-level expansions into a single super-kernel for each layer of the one-shot NAS supernet.  They reduce the NAS problem into finding the subset of kernel weights in each ConvNet layer. Fu \etal proposed AGD \cite{AGD} to compress GAN-based SR generator via NAS. Instead of searching for GAN architecture from scratch~\cite{gong2019autogan, gao2020adversarialnas}, AGD leverages knowledge from the pre-trained SR models to search for each cell's optimal operation (referred to as cell-level architecture search) and pruning of the input/output channels for each operation (\ie, kernel-level architecture search) via DARTS~\cite{DARTS}.

Yet, AGD overlooked the network-level search and kept the upsampling operations fixed at the network header part. Though the post-upsampling network design can be efficient, lack of network-level optimization might constrain its performance. On the contrary, Auto-DeepLab~\cite{Autodeeplab} proposed searching at the network level along with the cell-level structure, which forms a hierarchical architecture search space. Concretely, it models the network-level search with one trellis-like supernet, where the node represents intermediate feature maps, and arrows between nodes indicate the corresponding cell. The trellis-like design allows the method to explore for a general network-level search space. Likewise, Guo \etal \cite{Guo_2020} used hierarchical NAS idea for SR problem. The notable difference is that it leverages reinforcement learning for the cell-level and network-level architecture design. But, the method does not aim for kernel width pruning. Besides, the usage of reinforcement learning-based NAS adds practical challenges.

In the paper's remaining section, we first explain our trilevel search space followed by its continuous modeling and optimization, finally leading to experimental results.
\section{Trilevel Search Space and its Modeling}\label{sec:search_space}
We propose to search at kernel-level, cell-level, and network-level structures. We suggest a novel strategy for the network-level search space, where all candidate network paths are enumerated in a tree structure. For cell-level and kernel-level search, we leverage DARTS \cite{DARTS} and AGD \cite{AGD}.

\smallskip
\formattedparagraph{Network Level Search Space.}
Following AGD \cite{AGD} and SRResNet \cite{ledig2017photo},  we define our network-level search space. The reason for that is their supernet makes most of the computation in low-resolution feature space for computational efficiency. By fixing the network's stem (the first convolution layer) and header (the last convolution layer), we search for the path of the remaining five residual-in-residual (RiR) blocks and two upsampling blocks. We replace the dense blocks in RiR modules with five sequential layers containing searchable cell-level operators and kernel-level widths (see TrilevelNAS-A in Fig.\ref{fig:backbone}(a)). For efficient upsampling block design,  we replace the two upsampling blocks \cite{AGD} with one PixelShuffel block, which contains a convolution layer (with an output channel number being $n \times n \times3$) \cite{zhang2020aim}, and a PixelShuffel layer to reach the target resolution with an upscaling factor of $n$.  As the PixelShuffle block is fixed at the tail of the network, we add two regular convolution layers into the network-level search space to focus only on stacked 5 RiR blocks and 2 standard convolution layers for the network path search (see TrilevelNAS-B in Fig.\ref{fig:backbone}(b)).

\smallskip
\formattedparagraph{Network Level Search Space Modeling.} We avoid using a trellis-like structure to model the network-level search space because the trellis modeling aims to traverse all the network blocks' sequential paths. As shown in Fig.\ref{fig:tree-superpath}, each path starts from the first node and goes along a set of arrows to the target. Clearly, all the paths share most of the nodes and arrows. That leads to many redundant sharing among the network paths, the cells, and the kernels. Although the architecture sharing strategy saves training memory, it dramatically limits the search space. Further, the tight entanglement is likely to affect the learning on each path's contribution and the pruning of unnecessary paths. Consequently, we propose a tree structure for a flexible network-level path search modeling. As shown in Fig.\ref{fig:tree-superpath},  each node is merely connected to its father (if applicable) and children; hence the dependencies are highly relaxed. Nonetheless, we must maintain the associations at train time for lower memory consumption. Relaxing the correlations of distinct paths enables a flexible network-level search space. The lower dependency on paths may lead to a reliable association among cells and kernels due to their hierarchical connection. Moreover, the introduced tree modeling enables better disentanglement among the paths allowing us to perform pruning on redundant paths. Also, it supports the removal of the candidate paths once they contribute to the supernet.

\smallskip
\formattedparagraph{Cell Level and Kernel Level Search Space.} To define the search space at  cell-level and kernel-level, we followed AGD work \cite{AGD}. At cell-level, we search for five RiR blocks with each block containing five searchable cells, \ie, in total, 25 searchable cells. For each cell, we select one of the following candidate operations:

\begin{itemize}[leftmargin=*, topsep=0pt, noitemsep]
\small
    \item \emph{(i)} Conv 1$\times$1, \emph{(ii)} Residual Block (2 layers of Conv 3$\times$3 with a skip-connection), \emph{(iii)} Conv 3$\times$3, \emph{(iv)} Depthwise Block (Conv 1$\times$1 + Depthwise Conv3$\times$3 + Conv 1$\times$1).
\end{itemize}

\smallskip

\formattedparagraph{Cell Level and Kernel Level Search Space Modeling.} For cell-level search space modeling, we follow the DARTS cell modeling procedure. To model kernel-level search space, we follow AGD \cite{AGD} superkernel framework. For each convolution kernel, we set up a superkernel which has the full channels. To prune the channel number of the superkernel, a set of searchable expansion ratios $\phi=[\frac{1}{3}, \frac{1}{2}, \frac{4}{5}, \frac{5}{6}, 1]$ is defined, and parameters $\gamma_i$ which indicates the probability to choose the $i$-th expansion ratio is to be optimized. 
\section{Proposed Approach}
Our approach's key idea is to relax the discrete trilevel search space's explicit selection to an implicit selection from a hierarchical mixture of all the involved candidates in the search space. The continuous relaxation enables us to select the candidate with the highest contribution to the supernet in a fully-differentiable manner. To that end, we propose sparsestmax and its variant to perform continuous relaxation of the defined search space. Our modeling strategy can provide good sparsity and can seek dominant candidate architectures while preserving convexity and differentiability properties from softmax.

We introduce a novel sparsestmax ordering constraint at the network level for better path pruning. Ordering constraint makes the candidate paths to have heavier heads ( \ie, with more dominant contributions to the path) and lighter tails (\ie, with marginal contributions to the path) so that tails can be easily removed. Consequently, with the ordered sparsestmax on the mixture of all candidate networks, our supernet generally converges to a reasonably sparse network. As a benefit, the candidate architecture given by the sparsestmax activation is selected directly to design the optimal architecture. The discrepancy between the searched architecture and trained architecture gets highly decreased, enabling us to start training the network from the parameters learned during the search phase rather than from scratch. 

\smallskip
\formattedparagraph{Supernet Modeling with Sorted Sparsestmax.}
In our introduced network backbone \S \ref{sec:search_space}, since the upsampling layer is out of the network-level search space, each feature map (\ie, node) in the tree model can be used as the output of its corresponding path. Then the output can be fed into the upsampling layer to get the desired resolution. Hence, the aggregation over all the paths is reduced to the fusion over all the involved feature maps. In particular, we define a set of contribution weights $\beta$ for all the feature maps from the involved network paths based on our suggested tree model (Fig.\ref{fig:tree-superpath}). Therefore, the output of the supernet is a weighted combination of all the intermediate feature maps.  Given a tree model of the network-level search space with $N$ paths being $P=\{P_1, \ldots, P_N\}$ where $P_i$ has $M_i$ feature maps, the output of the whole supernet (\ie, the mixture of all the candidate paths) is defined as:
\begin{align}
    O_{tree}  = \sum_{i=0}^{N}\sum_{j=0}^{M_i}F_N(\beta_{i, j}; \bm{\beta}) f_{i, j}, 
    \label{tree_formulation}
\end{align}
where, $f_{i, j}$ is the feature map at the $j$-th layer of $P_i$, and $F_N(\beta_{i, j}; \bm{\beta})$ indicates the normalized combination weight over the feature map $f_{i,j}$. As discussed, instead of applying $\text{softmax}(\bm{\beta})$, we use the sparsestmax \cite{shao2019ssn} as follows to normalize the combination weights, which gives sparse contribution of the candidates to the mixture,
\begin{equation}
    F_{N}(\bm{\beta}, r):= \argmin_{\bm{q}\in \Delta_r^{k-1}} \|\bm{q}-\bm{\beta}\|_2^2,
    \label{eg:sparsestmax}
\end{equation}
where, $\Delta_r^{k-1}:=\{\bm{q} \in \mathbb{R}^K | \bm{1}^T\bm{q}=1, \|\bm{q} -\bm{u}\|_2 \geq  r, \bm{q} \geq \bm{0} \}$ indicates a simplex with a circular constraint $\bm{1}^T\bm{q}=1, \|\bm{q} -\bm{u}\|_2 \geq  r$, $\bm{u}=\frac{1}{K}\bm{1}$ is the center of the simplex, $\bm{1}$ is a vector of ones, and $r$ is radius of the circle. Sparsestmax returns the Euclidean projection of the input vector $\bm{\beta}$ onto the probability simplex. This projection is likely to touch the simplex boundary, in which case sparsestmax produces sparse distributions. To achieve a better sparsity, sparsestmax further introduces a circular constraint that enables a progressive production of sparsity by linearly increasing $r$ from zero to $r_c$, where $r_c$ is the radius of the circumcircle of the simplex. For detailed evaluations on the superiority of sparsestmax over softmax and sparsemax \cite{sparsemax,niculae2017regularized}, we refer readers to Shao \etal work \cite{shao2019ssn}.

While sparsestmax gives sparse distributions, it is not aligned well with our sequential setup. Given a path, $P_i$ that consists of $M_i$ nodes (\ie, feature maps), sparsestmax produces unordered non-zero combination weights on the nodes (extreme case is all of them distribute at the odd/even nodes). In that case, we cannot prune the path well unless the sparsity is ordered. We can perform network-level pruning provided that the non-zero combination weights are descending along the path so that all the zero weights appear at the tail of the path. Accordingly, we exploited sorted sparsestmax, which imposes an ordering constraint to the weights $\bm{\beta_i}$ within each path $P_i$.  It helps the output feature maps from shallower layers to share more contributions to the supernet. We formulate \textbf{sorted sparsestmax} as
\begin{equation}
\begin{aligned}
    F_{N}(\bm{\beta}, r):= \argmin_{\bm{q}\in \Delta_r^{k-1}} \|\bm{q}-\bm{\beta}\|_2^2  +\lambda\sum_{i}\sum_{j}(\beta_{i, j} - \beta_{i, j-1}) 
    \label{eg:sortedsparsestmax}
\end{aligned}
\end{equation}
here $\lambda$ is a trade-off constant.

\smallskip
\formattedparagraph{Supercell and Superkernel Modeling.} Since the cell-level network has no sequential properties, we apply sparsestmax directly to relax the discrete search space into a continuous one. We define a set of operations' contribution weights $\alpha$, such that the output is a normalized weighted combination of all the candidate operations. The output of $i^{th}$ cell $C_i$ is the weighted sum of the 4 features maps from the 4 candidates operations, \ie,
\begin{align}
    C_i = \sum_{o \in O}F_{N}(\alpha_i^{(o)}; \alpha_i)o(C_{i-1})
\end{align}
where $F_N$ corresponds to sparsestmax, $o(C_{i-1})$ is one operations selected from the cell-level space $O$ over $C_{i-1}$.

For the continuous relaxation of the kernel-level search space, one solution is to apply sparsestmax to combine all the involved kernels' expansions. However, as the number of kernels is huge, it can lead to exponentially large possible combinations. Therefore,  we adhere to apply the differentiable Gumbel-softmax sampling to keep the setup consistent with AGD \cite{AGD}. As for optimizing expansion ratio parameters $\gamma$, we follow AGD \cite{AGD}  sampling-based strategy. Mainly at each step of optimization, we sample only one expansion ratio based on $\gamma$ and train the model with the selected sub-kernel. Here, we introduced Gumbel-softmax~\cite{gumbel1954statistical} to approximate the sampling process. Thanks to the differentiable Gumbel-softmax, the ratio parameters $\gamma$ can be optimized in a differentiable framework.

\begin{algorithm}
	\footnotesize
	\caption{The Proposed TrileveNAS}\label{alg:trilevelnas}

	\begin{algorithmic}[1]
		\STATE \textbf{Input:} dataset $\chi=\{x_i\}^{N}_{i=1}$, pretrained generator $G_0$, search space and supernet $G$, epochs to pretrain ($T_1$), search ($T_2$) and train-from-scratch ($T_3$)
		\STATE \textbf{Output:} trained efficient generator $G^*$
		\STATE Equally split $\chi$ into $\chi_1$ and $\chi_2$ and initialize supernet weight $w$ and architecture parameters $\{\alpha,\beta,\gamma\}$ 
		with uniform distribution 
		\STATE \textit{\# First Step: Pretrain}
		\FOR{$t$ $\gets$ 1\ to\ $T_1$ } 
		\STATE Get a batch of data $X_1$ from $\chi_1$
		\FOR{$\gamma$ in [$\gamma_{\mathrm{max}}$, $\gamma_{\mathrm{min}}$, $\gamma_{\mathrm{random1}}$, $\gamma_{\mathrm{random2}}$]
		}
		\STATE  $g_w^{(t)} = \nabla_w d(G(X_1, \alpha,\beta, \gamma),G_0(X_1))$
		\STATE  $w^{(t+1)}$ = update($w^{(t)}$, $g_w^{(t)}$)
		\ENDFOR
		\ENDFOR
		
		\STATE \textit{\# Second Step: Search}
		\FOR{$t$ $\gets$ 1 \ to\ $T_2$ }
		\STATE Get a batch of data $X_1$ from $\chi_1$
		\STATE  $g_w^{(t)} = \nabla_w d(G(X_1, \alpha, \beta, \gamma),G_0(X_1))$
		\STATE  $w^{(t+1)}$ = update($w^{(t)}$, $g_w^{(t)}$)
		\STATE Get a batch of data $X_2$ from $\chi_2$
		\STATE  {\scriptsize $g_{\alpha}^{(t)} = \nabla_{\alpha} d(G(X_2, \alpha,\beta, \gamma),G_0(X_2)) + \lambda\omega_1 \nabla_{\alpha} F(\alpha|\beta,\gamma) $}
		\STATE  {\scriptsize $g_{\beta}^{(t)} = \nabla_{\alpha} d(G(X_2, \alpha,\beta, \gamma),G_0(X_2)) + \lambda\omega_1 \nabla_{\beta} F(\beta | \alpha,\gamma) $}
		\STATE  {\scriptsize $g_{\gamma}^{(t)} = \nabla_{\gamma} d(G(X_2, \alpha,\beta, \gamma),G_0(X_2)) + \lambda\omega_2 \nabla_{\gamma} F(\gamma|\alpha,\beta) $}
		\STATE  $\alpha^{(t+1)}$ = update($\alpha^{(t)}$, $g_{\alpha}^{(t)}$)
		\STATE  $\beta^{(t+1)}$ = update($\beta^{(t)}$, $g_{\beta}^{(t)}$)
		\STATE  $\gamma^{(t+1)}$ = update($\gamma^{(t)}$, $g_{\gamma}^{(t)}$)
		\ENDFOR
		
		\STATE \textit{\#{Third Step: Train from scratch}}
		\STATE Derive the searched architecture $G^*$ with maximal $\{\alpha,\beta,\gamma\}$ for each layer and re-initialize weight $w$.
		\FOR{$t$ $\gets$ 1\ to\ $T_3$ }
		\STATE Get a batch of data $X$ from $\chi$
		\STATE  $g_w^{(t)} = \nabla_w d(G^*(X),G_0(X))$
		\STATE  $w^{(t+1)}$ = update($w^{(t)}$, $g_w^{(t)}$)
		\ENDFOR
	\end{algorithmic}
\end{algorithm}

\begin{table*}
\footnotesize
    \centering
    \begin{threeparttable}[b]
    \begin{tabular}{cccccccc}
    \toprule
    \multirow{2}{*}{\textbf{Method}} &  \multirow{2}{*}{\textbf{Path}} & \textbf{Params} & \textbf{GFLOPS} & \multicolumn{2}{c}{\textbf{PSNR}}
    & \textbf{Search Cost}& \multirow{2}{*}{\textbf{Type}}\\
    \cline{5-6}
     &&\textbf{(M)}& (256$\times$256) & \textbf{Set5} & \textbf{Set14} & \textit{$\#$(GPUs)$\times$GPU Days}\\
     \midrule
     ESRGAN~\cite{ESRGAN} & -  & 16.70 &1176.6 & 30.44 & 26.28 & - & Manual \\
     ESRGAN-prune~\cite{DBLP:journals/corr/LiKDSG16} & -  & 1.6 & 113.1 & 28.07 & 25.21 & - & Manual\\
     SRGAN~\cite{SRGAN} & - & 1.52 & 166.7 & 29.40 & 26.02 & - & Manual\\ 
     \midrule
    TPSR~\cite{lee2020journey} &  - & \textbf{0.06}  & - & 29.60 & 26.88 & \color{red}{40$\times$12} & One-level NAS\\
     AGD~\cite{AGD}$^\dagger$ &  - & 0.56  & 117.7 & 30.36 & 27.21 & 1$\times$2 & Bi-level NAS\\
      AGD-AutoDeepLab &  [0,0,0,1,0,0,1] & 0.71 & 165.8 & 30.48 & 27.23 &  1$\times$4 & Tri-level NAS\\
    \midrule
     TrilevelNAS-A & [0,0,0,1,0,1] & \textbf{0.34} & 117.4 & 30.34 & 27.29 & 1$\times$8 & Tri-level NAS\\
     TrilevelNAS-B & [0,0,0]  & \textbf{0.24} & \textbf{15.4} & 29.80 & 27.06 & 1$\times$8 & Tri-level NAS \\
     \bottomrule
    \end{tabular}
\end{threeparttable}
    \caption{\footnotesize Quantitative results of visualization-oriented SR models with scaling factor 4. As for the listed Path results, we use `$0$' to indicate a \textit{RiR} block and `$1$' is a \textit{Upsampling} block/\textit{Conv} layer in TrilevelNAS-A/TrilevelNAS-B respectively.$^\dagger$ Reproduced AGD with official setup and implementation. The statistics clearly show that our method can supply a lighter model with manageable computational resources, and its PSNR performance favorable compares to the best methods. Hence, the proposed method performs better with all the evaluation metrics combined.} 
    
    \label{tab:visualization-quantitative-result}
\end{table*}

\begin{table*}
\footnotesize
    \centering
    \begin{threeparttable}[b]
    \begin{tabular}{ccccccc}
    \toprule
    \multirow{2}{*}{\textbf{Method}} &  \multirow{2}{*}{\textbf{Path}} & \textbf{Params} & \textbf{GFLOPS} & \multicolumn{2}{c}{\textbf{PSNR}} & \multirow{2}{*}{\textbf{Type}}\\
    \cline{5-6}
     &&\textbf{(M)}& 256$\times$256& \textbf{Set5} & \textbf{Set14}\\
     \midrule
     ESRGAN~\cite{ESRGAN} & -  & 16.70 &1176.6 & 32.70 & 28.95 & Manual \\
     \midrule
     HNAS$^*$ & Up: $9^{th}$ layer & 1.69 & 330.7 & 31.94 & 28.41 & Bi-level NAS\\
     AGD~\cite{AGD}$^{\ddagger}$ &  - & 0.90 & 140.2 & 31.85 & 28.40 & Bi-level NAS\\
      AGD-AutoDeepLab$^\dagger$ &  [0,0,0,1,0,0,1] & 0.71 & 165.8 & 31.83 & 28.38 & Tri-level NAS\\
    \midrule
     TrilevelNAS-B & [0, 0, 0, 0, 0, 1] & \textbf{0.51} & \textbf{33.3} & 31.62 & 28.26 & Tri-level NAS \\
     \bottomrule
    \end{tabular}
\end{threeparttable}
    \caption{\footnotesize Quantitative results of PSNR-oriented SR models with scaling factor 4. As for the listed Path results, we use `$0$' to indicate a \textit{RiR} block and `$1$' is a \textit{UpConv} layer. Here, the following symbols indicates * Reproduced HNAS for PSNR-oriented $\times$4 SR tasks, $^\ddagger$ Reproduced with AGD official setup and implementation, $^\dagger$ Transferred from the visualization-oriented model.
    Clearly, our method performs better with all the evaluation metrics combined.
    }
    \label{tab:psnr-quantitative-result}
\end{table*}

\smallskip
\formattedparagraph{Proxy Task and Optimization.}
For our trilevel NAS task, instead of training the model from scratch, we leverage the knowledge from pre-trained state-of-the-art image SR models via knowledge distillation \cite{AGD}. Concretely, we take a pre-trained generator from ESRGAN \cite{ESRGAN} as our teacher model. Therefore, the search phase's proxy task aims to search for a model $G$ by minimizing the knowledge distillation distance $d$ between the output from the model $G$ and that from the teacher model $G_0$. Besides, an efficient model is always favorable for the image SR task, so we consider involving the model efficiency term $H$ in our target. The training objective of the proposed TrilevelNAS is as follows
\begin{equation}
 \begin{aligned}
    \min_{G,\alpha, \beta, \gamma}\frac{1}{N}\sum_{i=0}^Nd(G(x_i; \alpha, \beta, \gamma), G_0(x_i))  + \lambda_f H(G; \alpha, \beta, \gamma)
    \label{full-loss}
\end{aligned}
\end{equation}
Here $\alpha, \beta, \gamma$ are the parameters for the continuous relaxation of the trilevel architecture search space, $d(G, G_0)$ is a distance for the image SR task, and $H$ is the computational budget based constraint on the searched architecture. Following AGD \cite{AGD}, we compute $d(G, G_0)$ with a combination of content loss $L_c$ (avoid color shift), perceptual loss $L_p$ (preserve visual and semantic details), and employ the model FLOPS as a measure of model efficiency.

Our continuous relaxation enables the optimization of the network parameters of $G$, and all the supernet architecture weights $\alpha, \beta, \gamma$. As the number of architecture parameters is much smaller than that of the network parameters, optimizing them jointly on one single training set is likely to overfit. So, we optimize the network weights and architecture parameters alternatively under the bi-level optimization framework \cite{DARTS}. We separate the dataset into a train set and validation set, where the network weights and architecture parameters are optimized on these two separate sets of data, respectively. Further, since the converged architecture and finally selected architecture is close due to sparsestmax-based supernet relaxation, we go for the new NAS training scheme (\ie, training from search) instead of the conventional training scheme (\ie, training from scratch). Pseudocode of our implementation is provided in Algorithm (\ref{alg:trilevelnas}).

\section{Experiments}
\begin{figure}[t]
\begin{center}
 \includegraphics[width=0.9\linewidth]{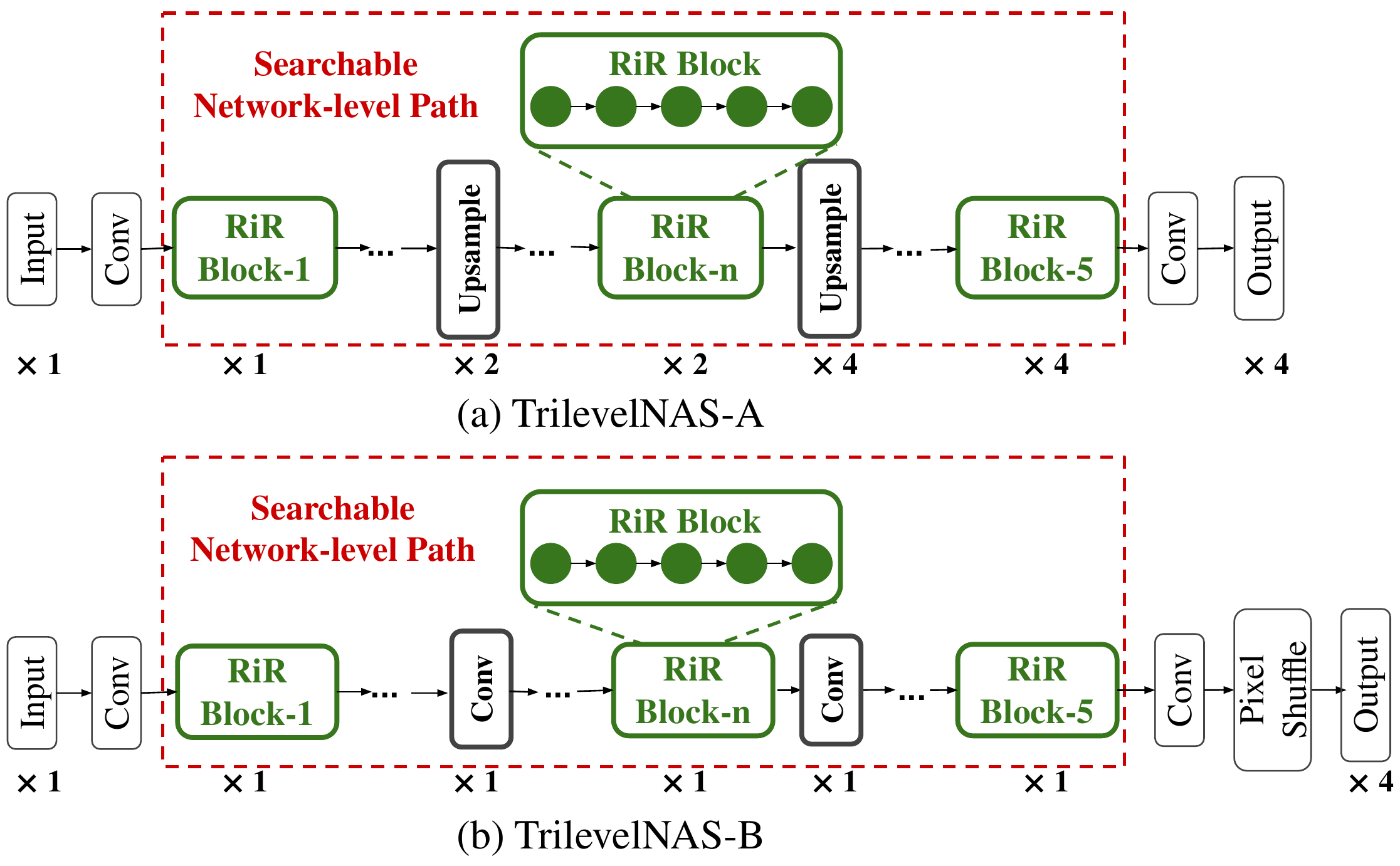}
\end{center}
\caption{\footnotesize (a) Image SR supernet backbone \textit{TrilevelNAS-A} with 5 RiR blocks and 2 upsampling layers. (b) Image SR supernet backbone \textit{TrilevelNAS-B} with 5 RiR blocks and a pixel shuffle layer.}
    \label{fig:backbone}
    \vspace{-0.3cm}
\end{figure}

 We performed search and training on DIV2K and Flickr2K datasets~\cite{timofte2017ntire}. Further, we evaluated the searched architectures on popular benchmarks including Set5~\cite{bevilacqua2012low}, Set14~\cite{zeyde2010single}, and DIV2K valid set. We used PSNR as a quantitative metric to report the results. Besides, we documented the model size and FLOPS (computed on 256 $\times$ 256 images) as quantitative metrics for model efficiency. We implemented the Network-Level search with two types of backbones: \textit{(i)} \textbf{TrilevelNAS-A:} Similar to AGD~\cite{AGD}, we search for a flexible path of 5 searchable RiR blocks and 2 upsampling blocks (see Fig.\ref{fig:backbone}(a)); \textit{(ii)} \textbf{TrilevelNAS-B:} For a more efficient SR model, we replace the 2 upsampling layers with a single PixelShuffle block, and search for the path of 5 RiR blocks and 2 convolution layers (see Fig.\ref{fig:backbone}(b)).

\begin{figure*}[t]
    \centering
    \includegraphics[width=0.9\linewidth]{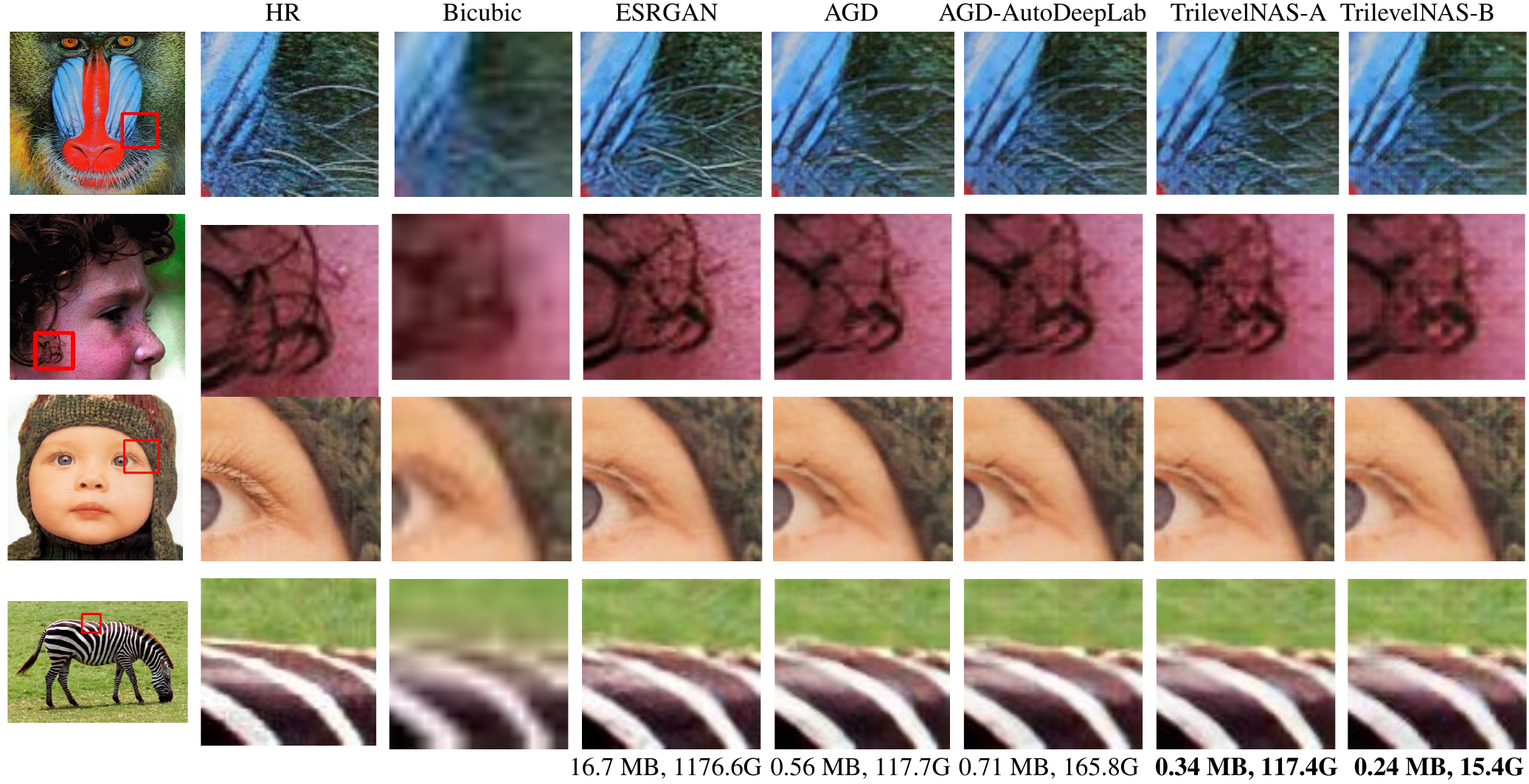}
    \caption{\footnotesize Visualization results of different SR models on Set5 and Set14. It can be observed that TrilevelNAS-B achieves comparable visual quality with very light model. At the bottom, we present the model parameter size (in MB) and FLOPS consumption (in GB) of different models respectively.}
    \label{fig:sample_results}
\end{figure*}

\smallskip
\formattedparagraph{Training Details.}
In differentiable NAS frameworks, optimization procedure generally consists of three steps: \textit{search}, \textit{model discretization} and \textit{train from scratch}. In contrast, due to the progressive sparsity property of sparsestmax, we propose a \textit{Train from Search} framework without the need for discretization step, which helps to reduce the model discrepancy between supernet and the discretized architecture. 
\begin{itemize}[leftmargin=*, topsep=0pt, noitemsep]

    \item \textit{Search:} Following AGD~\cite{AGD}, we split the search process into two phases: the pre-train phase and the search phase. As done by \cite{AGD}, in the pre-train phase, we use half of the training data to update the network weights for 100 epochs with only content loss $L_c$. Then in the search phase, we alternatively update the network weights and architecture weights for 100 epochs on two equally split training data. In this search stage, for the PSNR-oriented model search, we only optimize with the content loss $L_c$, and for visualization-oriented model search, we finetune with the perceptual loss for better visual quality.

    \item \textit{Train from Search:} As discussed, we do not need a model discretization step with a completely converged supernet after the entire search process. As a result, we can inherit the pre-trained network weights from the search stage as a good network initialization and continue to train the converged architecture. Following the same training process in AGD~\cite{AGD}, for the PSNR-oriented SR model, we train with only the content loss $L_c$ for 900 epochs. For the visualization-oriented SR model, we continue to finetune for 1800 epochs with a perceptual loss $L_p$\footnote{For more detailed experiment setups of search and training stages, please refer to our supplementary material.}.
\end{itemize}
\smallskip
\formattedparagraph{Visualization Oriented SR Model Search.}
We adopt ESRGAN~\cite{ESRGAN} model as the teacher to search for a visualization-oriented SR model. We compare our method to state-of-art SR GAN models (ESRGAN~\cite{ESRGAN}, a pruned ESRGAN baseline~\cite{DBLP:journals/corr/LiKDSG16} and SRGAN~\cite{SRGAN}) and NAS based visualization-oriented SR model (AGD~\cite{AGD} and TPSR~\cite{lee2020journey}). Note that we reproduce the AGD search and training results with their official code and default experiment setup. Besides, compared to our Tree-supernet-based Trilevel NAS, we apply the Trellis-supernet design of AutoDeepLab to the AGD backbone to implement Trilevel NAS for image SR task.

Table~\ref{tab:visualization-quantitative-result} show the statistical comparison of our approach against other competing methods. We additionally report the derived network path (`0' in path indicate the RiR blocks; `1' represent upsampling layers and convolution layers in TrilevelNAS-A and TrilevelNAS-B, respectively). Compared to the AutoDeepLab-based Trilevel NAS (AGD-AutoDeepLab), our TrellisNAS can prune RiR blocks or convolution layers, which significantly reduce model complexity. Without a significant change in performance, TrelevelNAS-A prunes one redundant RiR block and derives a much lighter model with a 0.34MB model size.  Even with smaller model size, TrilevelNAS-A still has comparable FLOPS against AGD. We notice that most of the FLOPS consumption comes from the blocks operating on high-dimension feature maps. Remarkably, our new backbone TrilevelNAS-B (see Fig.\ref{fig:backbone}(b)) has the potential to search for a flexible network-level path with both small model size and light GFLOPs. Compared to the original AGD and TrilevelNAS-A, the FLOPs consumption of TrilevelNAS-B is reduced by 4$\times$. Though TPSR derives a tiny architecture, its search cost (60$\times$ of our TrilevelNAS' search cost) is extremely expensive, and we also see a clear PSNR performance drop with this tiny model. We additionally compare the visual qualities of various SR models in Fig~\ref{fig:sample_results}, and we can see that our TrilevelNAS can derive more efficient SR models without performance loss in visual quality.

\smallskip
\formattedparagraph{PSNR Oriented SR Model Search.}
We implemented the PSNR-oriented SR model search on DIV2K and Flickr2K datasets and used the PSNR-oriented ESRGAN model as our teacher model. We have observed a clear FLOPs advantage of our new backbone (TrilevelNAS-B) over the original backbone (TrilevelNAS-A) in previous experiments. Here, we focus on the Trilevel search on our new backbone and aim for a more efficient SR model. We compare our derived model against our competitors in Table~\ref{tab:psnr-quantitative-result}. With comparable performance, the derived PSNR-oriented SR model on the new backbone prunes a convolution layer, and it is better in terms of the FLOPs than all other competitors.

Further, we compared our model with the winner and runner-up of the recent AIM challenge~\cite{zhang2020aim}. Following the challenge setup, we train and validate our model with 800 DIV2K training images and 100 valid images respectively\footnote{Please refer to supplementary materials for more training details}. Table~\ref{tab:challenge} provides the quantitative results of our method with competing methods, which include challenge winner (NJU\_MCG), runner-up (AiriA\_CG), and NAS-based SR models. It can be inferred from the statistics that our method, despite being significantly lighter, gives PSNR value comparable to competing approaches.

Note that the winner and runup methods train with extra Flickr2K dataset and take ground-truth HR images for strong supervision, which can be unrealistic for real-world application. Consequently, we adhere to use a teacher model for knowledge distillation. For that, we follow AGD, using the pretrained ESRGAN generator for the supervision. We can infer from Table~\ref{tab:challenge} that TrilevelNAS model being much lighter gives comparable results to the winning methods, which is desirable for mobile devices.

\smallskip
\formattedparagraph{Ablation Study.}
We conducted ablation studies to compare Sparsestmax against the Softmax, and understand the importance of our proposed Sorted Sparsestmax. Additionally, we study the effects of training under two different strategies \ie, train from scratch and train from search.

\noindent
\textit{(a) Softmax vs. Sparsestmax:} 
We study the supernet optimization with softmax and sparsestmax combination, respectively, keeping the same ordering constraint strategy. From Table~\ref{tab:ablation_softmax_sparsestmax}, we can observe that with weight ordering constraints, softmax and sparsestmax can prune one or two RiR blocks and have comparable performance in PSNR. However, sparsestmax converges to a more efficient model with a smaller model size and less flops consumption.

\begin{table}[]
\scriptsize
    \centering
    \begin{tabular}{cccccc}
    \toprule
    \multirow{2}{*}{\textbf{Method}} &  \textbf{Params} & \multirow{2}{*}{\textbf{GFLOPS}} & \textbf{PSNR} & \textbf{Extra} & \textbf{Ground}\\
     &\textbf{(M)}& & \textbf{[Val]} & \textbf{Data} & \textbf{Truth}\\
     \midrule
     NJU\_MCG & 0.43  & 27.10 & 29.04 & $\checkmark$ & $\checkmark$ \\
     AiriA\_CG & 0.687 & 44.98 & 29.00 & $\checkmark$ & $\checkmark$\\
     \midrule
     HNAS~\cite{Guo_2020}  & 1.69 & 330.74 & 28.86 & $\times$ & $\times$\\
     AGD~\cite{AGD} & 0.45 & 110.9 & 28.66 & $\times$ & $\times$\\
      AGD-AutoDeepLab & 0.71 & 165.8 & 28.83 & $\times$ & $\times$\\
    \midrule
   TrilevelNAS-B 
  & \textbf{0.27} & \textbf{17.33} & 28.52 & $\times$ & $\times$\\
     \bottomrule
    \end{tabular}
    \caption{\footnotesize Quantitative results on AIM 2020 Challenge~\cite{zhang2020aim}}
    \label{tab:challenge}
\end{table}

\begin{table}
\footnotesize
\centering
\resizebox{0.48\textwidth}{!}
{
    \begin{tabular}{cccccc}
    \toprule
    \multirow{2}{*}{\textbf{Method}} &  \multirow{2}{*}{\textbf{Path}} & \textbf{Params} & \multirow{2}{*}{\textbf{GFLOPS}} & \multicolumn{2}{c}{\textbf{PSNR}}\\
    \cline{5-6}
     &&\textbf{(M)}&& \textbf{Set5} & \textbf{Set14}\\
     \midrule
    Softmax & [0,0,1,0,1]  & 0.47 & 154.80 & 30.34 & 27.28 \\
    Sparsestmax & [0,0,0,1,0,1] & \textbf{0.34}  & \textbf{117.39}  & 30.34 & 27.29  \\
     \bottomrule
    \end{tabular}
}
    \caption{\footnotesize Quantitative results comparison of visualization-oriented SR models searched using softmax and sparsestmax supernet.
    }
    \label{tab:ablation_softmax_sparsestmax}
    \vspace{-0.3cm}
\end{table}

\begin{table}
\footnotesize
    \centering
    \begin{tabular}{cccccc}
    \toprule
    \multirow{2}{*}{\bm{$\lambda$}} &  \multirow{2}{*}{\textbf{Path}} & \textbf{Params} & \multirow{2}{*}{\textbf{GFLOPS}} & \multicolumn{2}{c}{\textbf{PSNR}}\\
    \cline{5-6}
     &&\textbf{(M)}&& \textbf{Set5} & \textbf{Set14}\\
     \midrule
     0 & [0,0,1,0,0,0,1]  & 0.58 & 184.08 & 30.32 & 27.21 \\
    0.01 & [0,0,1,0,0,1]  & 0.52 & 169.69 & 30.34 & 27.20 \\
    0.1 & [0,0,0,1,0,1] & 0.34  & 117.39  & 30.34 & 27.29  \\
     \bottomrule
    \end{tabular}
    \caption{\footnotesize Quantitative results of visualization-oriented SR models with scaling factor 4 for different ordering constraint strengths. As for the listed Path results, we use `$0$' to indicate a \textit{RiR} block and `$1$' is a \textit{UpConv} layer.}
    \label{tab:ablation_ordering_constraint}
\end{table}

\begin{figure}
    \centering
    \includegraphics[width=0.60\linewidth]{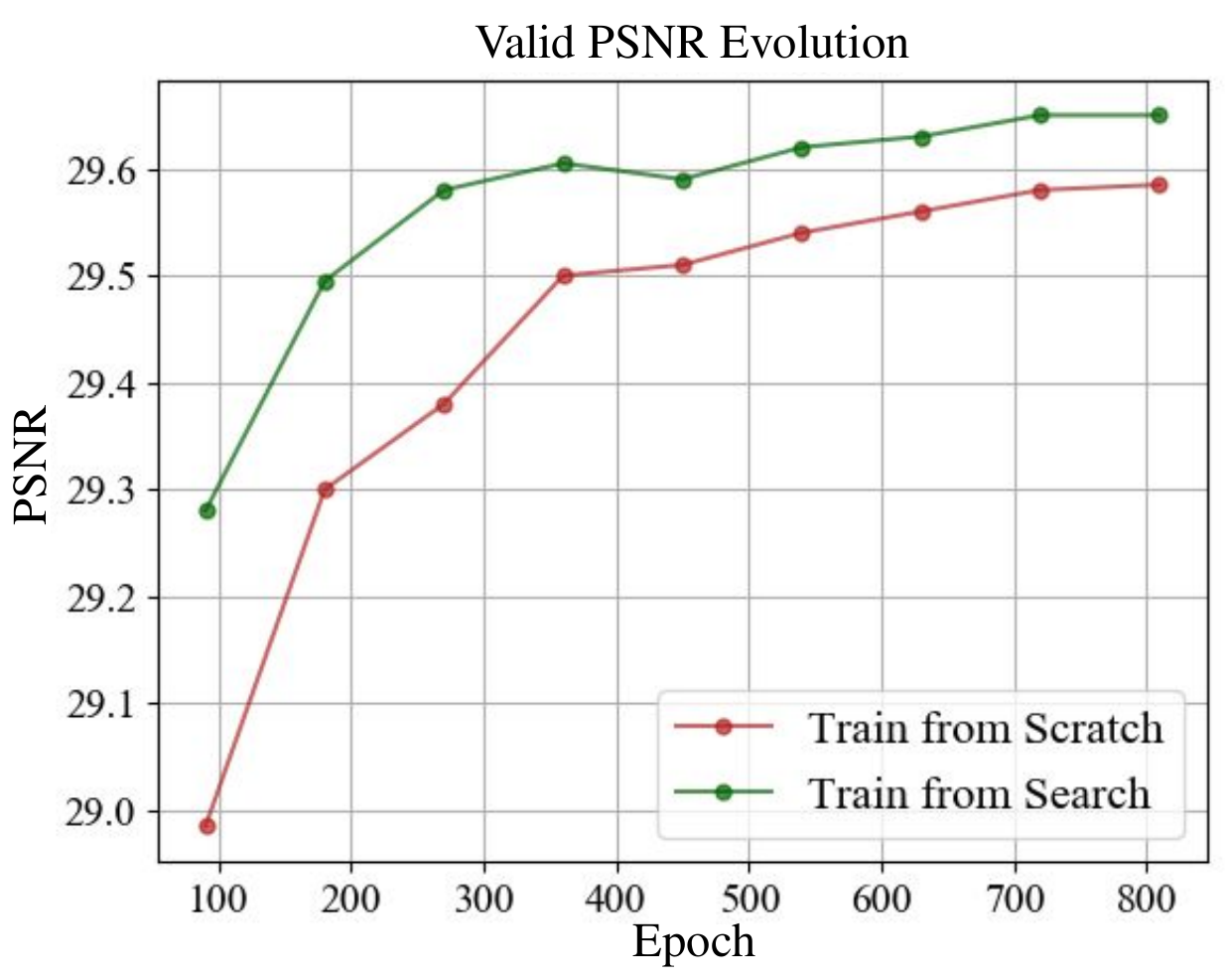}
    \caption{\footnotesize {Train from Search} vs {Train from Scratch} study. }
    \label{fig:single_two_phase_psnr}
    \vspace{-0.3cm}
\end{figure}
\noindent
\textit{(b) Sorted Sparsestmax:} We use a sorted sparsestmax combination of intermediate feature maps to encourage a shallow and efficient SR model without much loss in performance. In Eq.~\ref{eg:sortedsparsestmax} $\lambda$ controls the trade-off between the major constraint and the ordering constraint. A large ordering constraint tends to enforce a shallow network. We set $\lambda$ as 0, 0.01, 0.1, respectively, and examine its effects. Table~\ref{tab:ablation_ordering_constraint} show the derived paths and the corresponding model performance. We see that without a weight ordering constraint ($\lambda=0$), the TrellisNAS prefers a full network path with 5 RiR blocks and 2 upsampling blocks. When we impose ordering constraint with $\lambda$=0.01, 0.1, the tail RiR block is seen to be pruned, which yields more efficient SR models with small model size and FLOPs without loss in performance.

\noindent
\textit{(c) Train from Search vs. Train from Scratch:}
Lastly, we study the effect of inheriting weights from the search phase.  Fig.~\ref{fig:single_two_phase_psnr} shows the valid PSNR evolution of training from search and training from scratch, respectively. We can observe a clear advantage of inheriting weights from the search phase. With training from search strategy, we can converge to a better PSNR performance in fewer epochs.
\section{Conclusion}

\noindent
This paper presents a new trilevel NAS technique that provides lighter and efficient SR architecture with on-device computational resources, which compare favorably to the state-of-the-art PSNR results. The proposed NAS explores a more complete search space \ie, at network-level, cell-level, and  kernel-level. By introducing the proposed tri-level search space's continuous relaxation, we model the network path's hierarchical mixture.  Instead of relying on trellis-like network modeling, we show the utility of a tree-like supernet modeling with sparsestmax and sorted sparsestmax activation to the NAS framework.  The architecture obtained after our suggested optimization gives better SISR results ---considering all the vital evaluation metrics, hence supply a better alternative to the competing methods.

\section*{Acknowledgements}
This work was supported by the ETH Z\"urich Fund (OK), a Huawei (Moscow) project, an Amazon AWS grant, and an Nvidia GPU grant. Suryansh Kumar's work is supported by ``ETH Z\"urich Foundation and Google, Project Number: 2019-HE-323 (2)'' for bringing together best academic and industrial research.


\begin{thebibliography}{10}\itemsep=-1pt
	
	\bibitem{bevilacqua2012low}
	Marco Bevilacqua, Aline Roumy, Christine Guillemot, and Marie~Line
	Alberi-Morel.
	\newblock Low-complexity single-image super-resolution based on nonnegative
	neighbor embedding, 2012.
	
	\bibitem{proxylessnas}
	Han Cai, Ligeng Zhu, and Song Han.
	\newblock Proxylessnas: Direct neural architecture search on target task and
	hardware.
	\newblock {\em CoRR}, abs/1812.00332, 2018.
	
	\bibitem{chen2020progressive}
	Xin Chen, Lingxi Xie, Jun Wu, and Qi Tian.
	\newblock Progressive darts: Bridging the optimization gap for nas in the wild.
	\newblock 2020.
	
	\bibitem{chu2020fair}
	Xiangxiang Chu, Tianbao Zhou, Bo Zhang, and Jixiang Li.
	\newblock Fair darts: Eliminating unfair advantages in differentiable
	architecture search.
	\newblock In {\em ECCV}, 2020.
	
	\bibitem{SAN}
	Tao Dai, Jianrui Cai, Yongbing Zhang, Shu-Tao Xia, and Lei Zhang.
	\newblock Second-order attention network for single image super-resolution.
	\newblock In {\em Proceedings of the IEEE conference on computer vision and
		pattern recognition}, pages 11065--11074, 2019.
	
	\bibitem{SRCNN}
	Chao Dong, Chen~Change Loy, Kaiming He, and Xiaoou Tang.
	\newblock Image super-resolution using deep convolutional networks.
	\newblock {\em CoRR}, abs/1501.00092, 2015.
	
	\bibitem{AGD}
	Yonggan Fu, Wuyang Chen, Haotao Wang, Haoran Li, Yingyan Lin, and Zhangyang
	Wang.
	\newblock Autogan-distiller: Searching to compress generative adversarial
	networks.
	\newblock 2020.
	
	\bibitem{fu2020autogan}
	Yonggan Fu, Wuyang Chen, Haotao Wang, Haoran Li, Yingyan Lin, and Zhangyang
	Wang.
	\newblock Autogan-distiller: Searching to compress generative adversarial
	networks.
	\newblock {\em arXiv preprint arXiv:2006.08198}, 2020.
	
	\bibitem{gao2020adversarialnas}
	Chen Gao, Yunpeng Chen, Si Liu, Zhenxiong Tan, and Shuicheng Yan.
	\newblock Adversarialnas: Adversarial neural architecture search for gans.
	\newblock 2020.
	
	\bibitem{gong2019autogan}
	Xinyu Gong, Shiyu Chang, Yifan Jiang, and Zhangyang Wang.
	\newblock Autogan: Neural architecture search for generative adversarial
	networks.
	\newblock 2019.
	
	\bibitem{gumbel1954statistical}
	E.J. Gumbel.
	\newblock {\em Statistical Theory of Extreme Values and Some Practical
		Applications: A Series of Lectures}.
	\newblock Applied mathematics series. U.S. Government Printing Office, 1954.
	
	\bibitem{Guo_2020}
	Yong Guo, Yongsheng Luo, Zhenhao He, Jin Huang, and Jian Chen.
	\newblock Hierarchical neural architecture search for single image
	super-resolution.
	\newblock {\em IEEE Signal Processing Letters}, 27:1255–1259, 2020.
	
	\bibitem{DRCN}
	Jiwon Kim, Jung~Kwon Lee, and Kyoung~Mu Lee.
	\newblock Deeply-recursive convolutional network for image super-resolution.
	\newblock {\em CoRR}, abs/1511.04491, 2015.
	
	\bibitem{ledig2017photo}
	Christian Ledig, Lucas Theis, Ferenc Husz{\'a}r, Jose Caballero, Andrew
	Cunningham, Alejandro Acosta, Andrew Aitken, Alykhan Tejani, Johannes Totz,
	Zehan Wang, et~al.
	\newblock Photo-realistic single image super-resolution using a generative
	adversarial network.
	\newblock In {\em Proceedings of the IEEE conference on computer vision and
		pattern recognition}, pages 4681--4690, 2017.
	
	\bibitem{SRGAN}
	Christian Ledig, Lucas Theis, Ferenc Huszar, Jose Caballero, Andrew Cunningham,
	Alejandro Acosta, Andrew Aitken, Alykhan Tejani, Johannes Totz, Zehan Wang,
	and Wenzhe Shi.
	\newblock Photo-realistic single image super-resolution using a generative
	adversarial network.
	\newblock 2017.
	
	\bibitem{lee2020journey}
	Royson Lee, {\L}ukasz Dudziak, Mohamed Abdelfattah, Stylianos~I Venieris, Hyeji
	Kim, Hongkai Wen, and Nicholas~D Lane.
	\newblock Journey towards tiny perceptual super-resolution.
	\newblock In {\em European Conference on Computer Vision}, pages 85--102.
	Springer, 2020.
	
	\bibitem{DBLP:journals/corr/LiKDSG16}
	Hao Li, Asim Kadav, Igor Durdanovic, Hanan Samet, and Hans~Peter Graf.
	\newblock Pruning filters for efficient convnets.
	\newblock {\em CoRR}, abs/1608.08710, 2016.
	
	\bibitem{9157494}
	M. {Li}, J. {Lin}, Y. {Ding}, Z. {Liu}, J.~Y. {Zhu}, and S. {Han}.
	\newblock Gan compression: Efficient architectures for interactive conditional
	gans.
	\newblock In {\em 2020 IEEE/CVF Conference on Computer Vision and Pattern
		Recognition (CVPR)}, pages 5283--5293, 2020.
	
	\bibitem{SRFBN}
	Zhen Li, Jinglei Yang, Zheng Liu, Xiaomin Yang, Gwanggil Jeon, and Wei Wu.
	\newblock Feedback network for image super-resolution.
	\newblock {\em CoRR}, abs/1903.09814, 2019.
	
	\bibitem{Autodeeplab}
	Chenxi Liu, Liang{-}Chieh Chen, Florian Schroff, Hartwig Adam, Wei Hua, Alan~L.
	Yuille, and Li Fei{-}Fei.
	\newblock Auto-deeplab: Hierarchical neural architecture search for semantic
	image segmentation.
	\newblock {\em CoRR}, abs/1901.02985, 2019.
	
	\bibitem{DARTS}
	Hanxiao Liu, Karen Simonyan, and Yiming Yang.
	\newblock Darts: Differentiable architecture search.
	\newblock {\em arXiv preprint arXiv:1806.09055}, 2018.
	
	\bibitem{sparsemax}
	Andr{\'{e}} F.~T. Martins and Ram{\'{o}}n~Fernandez Astudillo.
	\newblock From softmax to sparsemax: {A} sparse model of attention and
	multi-label classification.
	\newblock {\em CoRR}, abs/1602.02068, 2016.
	
	\bibitem{niculae2017regularized}
	Vlad Niculae and Mathieu Blondel.
	\newblock A regularized framework for sparse and structured neural attention.
	\newblock In {\em Advances in neural information processing systems}, pages
	3338--3348, 2017.
	
	\bibitem{real2019regularized}
	Esteban Real, Alok Aggarwal, Yanping Huang, and Quoc~V Le.
	\newblock Regularized evolution for image classifier architecture search.
	\newblock 2019.
	
	\bibitem{real2017large}
	Esteban Real, Sherry Moore, Andrew Selle, Saurabh Saxena, Yutaka~Leon Suematsu,
	Jie Tan, Quoc~V Le, and Alexey Kurakin.
	\newblock Large-scale evolution of image classifiers.
	\newblock In {\em International Conference on Machine Learning}, pages
	2902--2911, 2017.
	
	\bibitem{sparsestmax}
	Wenqi Shao, Tianjian Meng, Jingyu Li, Ruimao Zhang, Yudian Li, Xiaogang Wang,
	and Ping Luo.
	\newblock {SSN:} learning sparse switchable normalization via sparsestmax.
	\newblock {\em CoRR}, abs/1903.03793, 2019.
	
	\bibitem{shao2019ssn}
	Wenqi Shao, Tianjian Meng, Jingyu Li, Ruimao Zhang, Yudian Li, Xiaogang Wang,
	and Ping Luo.
	\newblock Ssn: Learning sparse switchable normalization via sparsestmax.
	\newblock 2019.
	
	\bibitem{stamoulis2019single}
	Dimitrios Stamoulis, Ruizhou Ding, Di Wang, Dimitrios Lymberopoulos, Bodhi
	Priyantha, Jie Liu, and Diana Marculescu.
	\newblock Single-path nas: Designing hardware-efficient convnets in less than 4
	hours.
	\newblock In {\em Joint European Conference on Machine Learning and Knowledge
		Discovery in Databases}, pages 481--497. Springer, 2019.
	
	\bibitem{timofte2017ntire}
	Radu Timofte, Eirikur Agustsson, Luc Van~Gool, Ming-Hsuan Yang, and Lei Zhang.
	\newblock Ntire 2017 challenge on single image super-resolution: Methods and
	results.
	\newblock In {\em Proceedings of the IEEE conference on computer vision and
		pattern recognition workshops}, pages 114--125, 2017.
	
	\bibitem{SRDenseNet}
	Tong Tong, Gen Li, Xiejie Liu, and Qinquan Gao.
	\newblock Image super-resolution using dense skip connections.
	\newblock pages 4809--4817, 10 2017.
	
	\bibitem{ESRGAN}
	Xintao Wang, Ke Yu, Shixiang Wu, Jinjin Gu, Yihao Liu, Chao Dong, Chen~Change
	Loy, Yu Qiao, and Xiaoou Tang.
	\newblock {ESRGAN:} enhanced super-resolution generative adversarial networks.
	\newblock {\em CoRR}, abs/1809.00219, 2018.
	
	\bibitem{zeyde2010single}
	Roman Zeyde, Michael Elad, and Matan Protter.
	\newblock On single image scale-up using sparse-representations.
	\newblock In {\em International conference on curves and surfaces}, pages
	711--730. Springer, 2010.
	
	\bibitem{zhang2020memoryefficient}
	Haokui Zhang, Ying Li, Hao Chen, and Chunhua Shen.
	\newblock Memory-efficient hierarchical neural architecture search for image
	denoising.
	\newblock 2020.
	
	\bibitem{zhang2020aim}
	Kai Zhang, Martin Danelljan, Yawei Li, Radu Timofte, Jie Liu, Jie Tang,
	Gangshan Wu, Yu Zhu, Xiangyu He, Wenjie Xu, et~al.
	\newblock Aim 2020 challenge on efficient super-resolution: Methods and
	results.
	\newblock {\em arXiv preprint arXiv:2009.06943}, 2020.
	
	\bibitem{zoph2017neural}
	Barret Zoph and Quoc~V Le.
	\newblock Neural architecture search with reinforcement learning.
	\newblock {\em ICLR}, 2017.
	
	\bibitem{zoph2018learning}
	Barret Zoph, Vijay Vasudevan, Jonathon Shlens, and Quoc~V Le.
	\newblock Learning transferable architectures for scalable image recognition.
	\newblock In {\em Proceedings of the IEEE conference on computer vision and
		pattern recognition}, pages 8697--8710, 2018.
	
\end{thebibliography}

\newpage

\section*{Supplementary Material}
\appendix


In the material, we first provide a detailed description of our trilevel NAS algorithm's coding implementation outlined in the main paper, and the coding platform details, such as hardware and software used in our implementation are provided. A critical ablation study is performed and discussed to show the importance of trilevel optimization over bilevel optimation with sparsestmax constraint. Further, concrete information on the optimal architecture obtained after trilevel optimization is tabulated. Finally, we supply some more visualization results of our super-resolved images, which shows our approach's efficiency.

\section{Implementation Details}
In this section, we describe the implementation details of our trilevel NAS algorithm for the SR task. The search strategy using visualization and the PSNR-oriented teacher model is discussed, followed by the training strategy. The code is developed with Python 3.6 and Pytorch 1.3.1. The complete search process takes 8 GPU days on single NVIDIA Tesla V100 (16GB RAM) and the training of a PSNR-oriented and visualization-oriented SR model take 0.5 and 1.5 GPU days respectively.

\subsection{Search}
\noindent
We implement our search phase using DIV2K and Flickr2K dataset \cite{timofte2017ntire}.

\noindent
\textbf{(a) Visualization-oriented:} 
Here,  we took the visualization-oriented ESRGAN model as a teacher model for the SR model search task. Subsequently, we perform the search task in two phase: \textit{(a) Pretrain}: using half of the training data, we optimize the content loss $L_c$ and train the supernet weights without updating architecture parameters. We train for 100 epochs using Adam optimizer, and at each optimization step, 3 patches of size 32$\times$32 are randomly cropped. For architecture parameters, we use a constant learning rate $3\times10^{-4}$, and for network weights, we set the initial learning rate as $1\times10^{-4}$, decayed by 0.5 at $25^\textrm{th}$, $50^\textrm{th}$, $75^\textrm{th}$ epoch. \textit{(b) Search}: optimizing content loss $L_c$ and perceptual loss $L_p$, we alternatively update the network weights and architecture weights for 100 epochs on two equally split training data. The optimizer follows the same setting as in the pretrain phase.

\noindent
\textbf{(b) PSNR-oriented:} 
For this, we took the PSNR-oriented ESRGAN model as teacher model for SR model search and followed the similar procedure \ie, \textit{Pretrain} and \textit{Search}, and however, here we optimize only the content loss $L_c$ in the two phases.

\subsection{Train}
\noindent
We inherit the weights from supernet and train with DIV2K and Flickr2K datasets \cite{timofte2017ntire} in the training phase. Patches of size $32\times32$ are randomly cropped, and the batch size is set to 16.

\noindent
\textbf{Visualization-oriented:} 
SR model training is conducted in two steps: \textit{(a) Pretrain} by minimizing content loss $L_c$ for 900 epochs. Adam optimizer is used with initial learning rate $1\times10^{-4}$ and learning rate decays by 0.5 at $225^\textrm{th}$, $450^\textrm{th}$, $675^\textrm{th}$ epoch. \textit{(b) Fine-tune} with perceptual loss $L_p$ for 1800 epochs. We also train with Adam optimizer with initial learning rate $1\times10^{-4}$, decayed by 0.5 at $225^\textrm{th}$, $450^\textrm{th}$, $900^\textrm{th}$, $1350^\textrm{th}$ epoch.

\noindent
\textbf{PSNR-oriented:} SR model is trained by minimizing the content loss $L_c$ for 900 epochs, and the optimizer follows the same setting as the pretrain phase in visualization-oriented model training. Note that, for the challenge setup, we only train our proposed model with DIV2K (without using extra data), and merely use a teacher for data distillation instead of using the ground truth for strong supervision, which might lead to marginally inferior performance. We first crop LR images in DIV2K into sub-images of size $120\times120$. We use Adam to train for 300 epochs, where the learning rate decays by 0.5 at $75^\textrm{th}$, $150^\textrm{th}$, $225^\textrm{th}$ epoch.

\begin{figure*}
	\centering
	\includegraphics[width=0.90\linewidth]{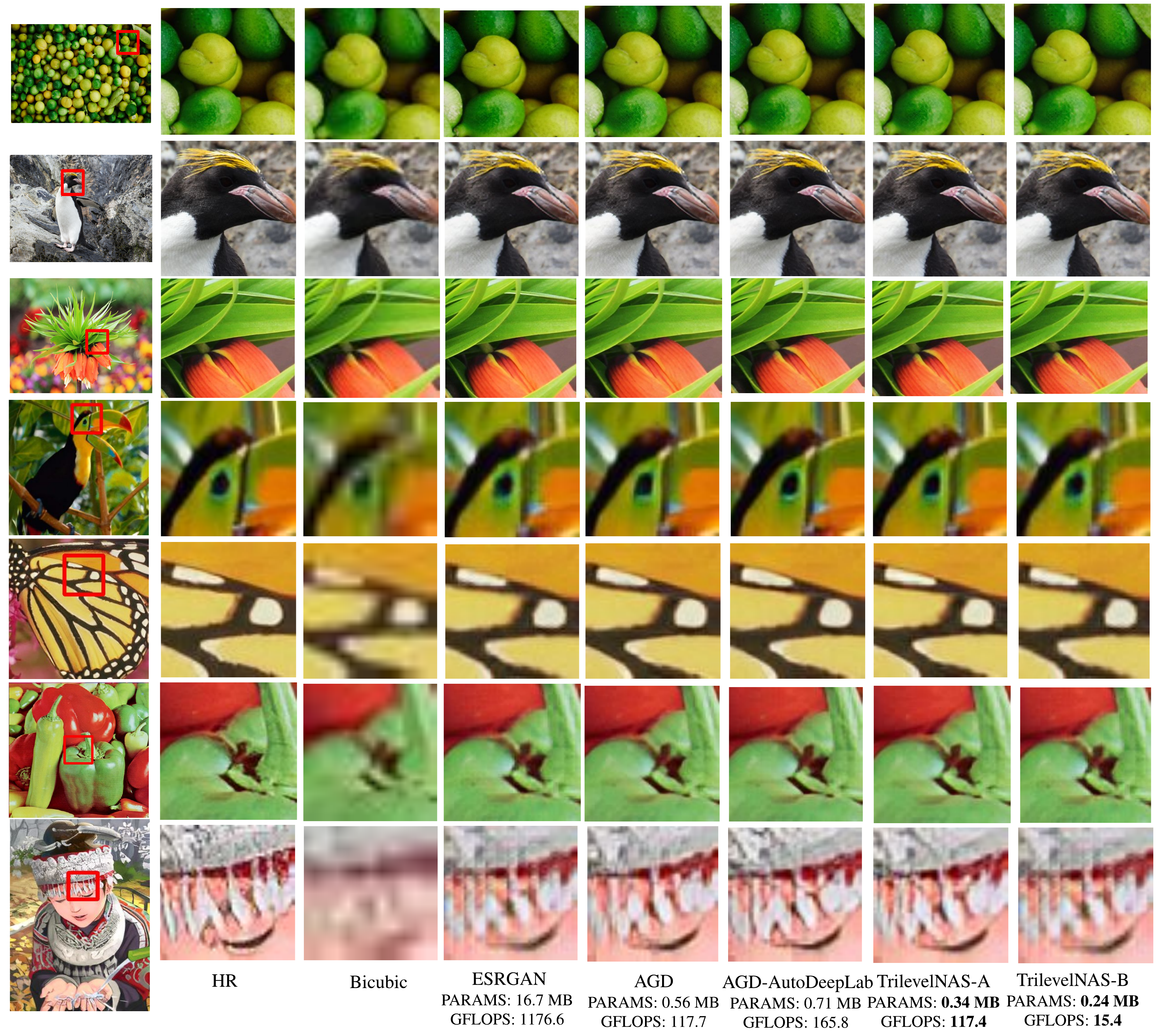}
	\caption{Visualization results of different SR models on Set5, Set14 and DIV2K valid set. Our TrilevelNAS-A and TrilevelNAS-B achieve comparable visual quality
		with very light model.}
	\label{fig:supp_visual_results}
\end{figure*}

\begin{table}
	\footnotesize
	\centering
	\resizebox{0.48\textwidth}{!}
	{
		\begin{tabular}{cccccc}
			\toprule
			\multirow{2}{*}{\textbf{Method}} &  \multirow{2}{*}{\textbf{Path}} & \textbf{Params} & \multirow{2}{*}{\textbf{GFLOPS}} & \multicolumn{2}{c}{\textbf{PSNR}}\\
			\cline{5-6}
			&&\textbf{(M)}&& \textbf{Set5} & \textbf{Set14}\\
			\midrule
			BilevelNAS & [0,0,0,0,0,1,1]$^*$  & 0.52 & 115.39 & 30.32 & 27.22 \\
			TrilevelNAS & [0,0,0,1,0,1] & \textbf{0.34}  & 117.39  & 30.34 & 27.29\\
			\bottomrule
		\end{tabular}
	}
	\caption{\footnotesize Quantitative results comparison of BilevelNAS (kernel- and cell-level) and TrilevelNAS (kernel-, cell- and network-level). Here $*$ indicates that the path is fixed rather than being searched.
	}
	\label{tab:ablation_bilevel_trilevel}
\end{table}

\begin{table*}
	\small
	\centering
	\begin{minipage}{0.90\linewidth}
		\begin{tabular}{c|ccccc}
			\toprule
			\textbf{Path} & \multicolumn{5}{c}{\textbf{[0, 0, 0, 1, 0, 1]}}\\
			\toprule
			RiR Block ID& OP1 & OP2 & OP3 & OP4 & OP5 \\
			\midrule
			0  & (Conv $1 \times 1$, 64) & (Conv $3 \times 3$, 24) & (Conv $1 \times 1$, 24) & (Conv $1 \times 1$, 24) & (Conv $3 \times 3$, 64) \\
			1 & (DwsBlock, 32)  & (DwsBlock, 32)  & (DwsBlock, 64)  & (Conv $3 \times 3$, 24) &  (ResBlock, 64) \\
			2 & (Conv $1 \times 1$, 64)  & (Conv $1 \times 1$, 64)  & (Conv $3 \times 3$, 40) & (Conv $1 \times 1$, 32) &  (DwsBlock, 64) \\
			3 & (Conv $3 \times 3$, 24)  & (Conv $3 \times 3$, 64)  & (Conv $1 \times 1$, 32) & (Conv $3 \times 3$, 32) &  (Conv $3 \times 3$, 64) \\
			\bottomrule
		\end{tabular}
		\caption{\small Visualization-oriented SR model architecture searched on DIV2K and Flickr2K dataset~\cite{timofte2017ntire} with original AGD backbone (TrilevelNAS-A). We present the network-level architecture (i.e., path of stacking RiR blocks and upsampling layers), the cell-level structures (i.e., the selection of operations in OP1-OP5 in each RiR block) and the kernel-level structures (i.e., the selection of output channels from the full 64 channels). In the path, '0' indicates a RiR block and the '1' indicates an upsamling layer. For each OP, the format $(A, B)$ indicates selecting the operation $A$ with its kernel width being $B$.}
		\label{tab:visual-A}
		\vspace{0.5cm}
		\begin{tabular}{c|ccccc}
			\toprule
			\textbf{Path} & \multicolumn{5}{c}{\textbf{[0, 0, 0]}}\\
			\toprule
			RiR Block ID& OP1 & OP2 & OP3 & OP4 & OP5 \\
			\midrule
			0  & (DwsBlock, 32) & (Conv $1 \times 1$, 32) & (DwsBlock, 24) & (DwsBlock, 40) & (DwsBlock, 64) \\
			1 & (DwsBlock, 24)  & (Conv $1 \times 1$, 56)  & (Conv $3 \times 3$, 24)  & (DwsBlock, 24) &  (Conv $3 \times 3$, 64) \\
			2 & (ResBlock, 64)  & (Conv $1 \times 1$, 24)  & (Conv $1 \times 1$, 24) & (Conv $1 \times 1$, 24) &  (Conv $1 \times 1$, 64) \\
			\bottomrule
		\end{tabular}
		\caption{\small Visualization-oriented SR model architecture searched on DIV2K and Flickr2K dataset~\cite{timofte2017ntire} with our proposed new backbone (TrilevelNAS-B). We present the network-level architecture  (i.e., path of stacking RiR blocks and convolutional layers), the cell-level structures (i.e., the selection of operations in OP1-OP5 in each RiR block) and the kernel-level structures (i.e., the selection of output channels from the full 64 channels). In the path, '0' indicates a RiR block and the '1' indicates a convolutional layer. For each OP, the format $(A, B)$ indicates selecting the operation $A$ with its kernel width being $B$.}
		\label{tab:visual-B}
		\vspace{0.5cm}
		\centering
		\begin{tabular}{c|ccccc}
			\toprule
			\textbf{Path} & \multicolumn{5}{c}{\textbf{[0, 0, 0, 0, 0, 1]}}\\
			\toprule
			RiR Block ID& OP1 & OP2 & OP3 & OP4 & OP5 \\
			\midrule
			0  & (Conv $1 \times 1$, 32) & (Conv $1 \times 1$, 24) & (DwsBlock, 64) & (DwsBlock, 24) & (Conv $1 \times 1$, 64) \\
			1 & (DwsBlock, 32)  & (ResBlock, 40)  & (DwsBlock, 56)  & (Conv $3 \times 3$, 40) &  (Conv $3 \times 3$, 64) \\
			2 & (Conv $3 \times 3$, 64)  & (Conv $1 \times 1$, 64)  & (Conv $3 \times 3$, 24) & (Conv $1 \times 1$, 32) &  (DwsBlock, 64) \\
			3 & (DwsBlock, 64)  & (DwsBlock, 56)  & (Conv $3 \times 3$, 24) & (Conv $3 \times 3$, 24) &  (DwsBlock, 64) \\
			4 & (DwsBlock, 32)  & (Conv $3 \times 3$, 64)  & (Conv $3 \times 3$, 64) & (Conv $3 \times 3$, 24) &  (Conv $3 \times 3$, 64) \\
			\bottomrule
		\end{tabular}
		\caption{\small PSNR-oriented SR model architecture searched on DIV2K and Flickr2K~\cite{timofte2017ntire} with our new backbone (TrilevelNAS-B). We present the network-level architecture  (i.e., path of stacking RiR blocks and convolutional layers), the cell-level structures (i.e., the selection of operations in OP1-OP5 in each RiR block) and the kernel-level structures (i.e., the selection of output channels from the full 64 channels). In the path, '0' indicates a RiR block and the '1' indicates a convolutional layer. For each OP, the format $(A, B)$ indicates selecting the operation $A$ with its kernel width being $B$.}
		\label{tab:psnr-B}
		\vspace{0.5cm}
		\centering
		\begin{tabular}{c|ccccc}
			\toprule
			\textbf{Path} & \multicolumn{5}{c}{\textbf{[0, 0, 1, 1]}}\\
			\toprule
			RiR Block ID& OP1 & OP2 & OP3 & OP4 & OP5 \\
			\midrule
			0  & (Conv $1 \times 1$, 40) & (Conv $1 \times 1$, 24) & (Conv $1 \times 1$, 56) & (ResBlock, 56) & (Conv $1 \times 1$, 64) \\
			1 & (DwsBlock, 64)  & (DwsBlock, 56)  & (Conv $3 \times 3$, 32)  & (Conv $3 \times 3$, 24) &  (Conv $1 \times 1$, 64) \\
			\bottomrule
		\end{tabular}
		\caption{\small PSNR-oriented SR model architecture searched on DIV2K and Flickr2K~\cite{timofte2017ntire} with our proposed TrilevelNAS-B on AIM challenge~\cite{zhang2020aim}. We present the network-level structures (i.e., path of stacking RiR blocks and convolutional layers), the cell-level structures (i.e., the selection of operations in OP1-OP5 in each RiR block) and the kernel-level structure (i.e., the selection of output channels from the full 64 channels). In the path, '0' indicates a RiR block and the '1' indicates a convolutional layer. For each OP, the format $(A, B)$ indicates selecting the operation $A$ with its kernel width being $B$.}
		\label{tab:psnr-AIM}
	\end{minipage}
\end{table*}

\section{Ablation Study}
\noindent
We performed the following ablation study the advantage of performing trilevel NAS as compared to bilevel NAS \cite{fu2020autogan} \cite{Autodeeplab} using the sparsestmax continuous relaxation \cite{shao2019ssn}. This experimental setup shall help understand the utility of exhaustive search to obtain the light-weight model without sacrificing much on the performance.

\noindent
\textbf{Bilevel vs. Trilevel:}  To show the necessity and advantage of introducing our network-level search in addition to cell-level and kernel-level search, we compare the BilevelNAS, which has a fixed network-level design (\ie, path: [0, 0, 0, 0, 0, 1, 1]) with our TrilevelNAS. For BilevelNAS, we use AGD~\cite{fu2020autogan}, and for a fair comparison, we replace the original softmax combination in AGD with sparsestmax combination. From Table~\ref{tab:ablation_bilevel_trilevel}, we can conclude that the additional network-level search in TrilevelNAS enables us to derive a lighter model with comparable performance. It can be inferred from the Table~\ref{tab:ablation_bilevel_trilevel} statistics that the number of parameters in our approach is significantly smaller as compared to BilevelNAS, whereas the PSNR value is slightly better with comparable GFLOPS on Set5 and Set14.

\section{Architecture Details}
Table~\ref{tab:visual-A}-\ref{tab:psnr-AIM} present our derived network-level (path), cell-level (5 searchable operations in each RiR block) and kernel-level (output channel number) architectures. Table \ref{tab:visual-A}-\ref{tab:visual-B} show the SR  architecture model obtained using visualization-oriented search strategy on DIV2K and Flick2K dataset, whereas, Table \ref{tab:psnr-B}-\ref{tab:psnr-AIM} provide the SR architecture obtained using PSNR-oriented search strategy.

The obtained network-level path is shown in first row of the respective table. The entry `$\mathbf{0}$' and `$\mathbf{1}$' in the path-index row-vector indicates RiR block and upsampling/convolution layer respectively. In addition, we lists the cell-level structures (i.e., the selection of operations in OP1-OP5 in each RiR block) and the kernel-level structures (i.e., the selection of output channels from the full 64 channels) in Table \ref{tab:psnr-B}-\ref{tab:psnr-AIM}. For each OP, the format $(A, B)$ indicates selecting the operation $A$ with its kernel width being $B$. Regarding the types of different OPs, DwsBlock, ResBlock, Conv symbolises depthwise convolution block, residual block and convolution block respectively.


\section{More Visualization Results}
Figure \ref{fig:supp_visual_results} show the visual comparison of our approach against other competing methods. The last two columns in the figure show the results obtained using our approach with TrilevelNAS-A and TrilevelNAS-B search strategy, respectively. Clearly, our method supplies a significantly lighter model and provides a super-resolved image that is perceptually as good as, if not better, than other approach results.

\end{document}